\def\year{2020}\relax
\documentclass[letterpaper]{article} 
\usepackage{aaai20}  
\usepackage{times}  
\usepackage{helvet} 
\usepackage{courier}  
\usepackage{booktabs}
\usepackage[hyphens]{url}  
\usepackage{graphicx} 
\usepackage{amssymb}
\usepackage{multirow}
\usepackage{graphicx}  
\usepackage[ruled,linesnumbered]{algorithm2e}
\nocopyright

\title{Decompressing Knowledge Graph Representations for Link Prediction}

\author{Xiang Kong\thanks{Equal contribution, in alphabetical order.}, Xianyang Chen\footnotemark[1], Eduard Hovy\\
  Language Technologies Institute\\
  Carnegie Mellon University\\
  \{\tt  xiangk, xianyangc, hovy\}@cs.cmu.edu}

\begin{document}

\maketitle

\begin{abstract}
This paper studies the problem of predicting missing relationships between entities in knowledge graphs through learning their representations. Currently, the majority of existing link prediction models employ simple but intuitive scoring functions and relatively small embedding size so that they could be applied to large-scale knowledge graphs. However, these properties also restrict the ability to learn more expressive and robust features. Therefore, diverging from most of the prior works which focus on designing new objective functions, we propose, DeCom, a simple but effective mechanism to boost the performance of existing link predictors such as DistMult, ComplEx, etc, through extracting more expressive features while preventing overfitting by adding just a few extra parameters.
Specifically, embeddings of entities and relationships are first decompressed to a more expressive and robust space by decompressing functions, then knowledge graph embedding models are trained in this new feature space. Experimental results on several benchmark knowledge graphs and advanced link prediction systems demonstrate the generalization and effectiveness of our method. Especially, RESCAL + DeCom achieves state-of-the-art performance on the FB15k-237 benchmark across all evaluation metrics.
In addition, we also show that compared with DeCom, explicitly increasing the embedding size significantly increase the number of parameters but could not achieve promising performance improvement. Code has been released\footnote{\url{https://github.com/shawnkx/Decom}}.
\end{abstract}

\section{Introduction}
Recently, knowledge bases (KBs) such as Freebase~\cite{bollacker2008freebase}, WordNet~\cite{miller1995wordnet}, Yago~\cite{suchanek2007yago} has proven useful in many tasks, including reading comprehension, recommendation system, information retrieval, etc. These KBs collect facts related to the real world as directed graphs (knowledge graphs), in which entities (nodes) are connected by their relationships (edges). As a result, a fact is represented by a triple ($\mathbf{s}$, $\mathbf{r}$, $\mathbf{o}$), i.e., a relation $\mathbf{r}$ between a subject entity $\mathbf{s}$ and an object entity $\mathbf{o}$. 

Although these knowledge graphs contain millions of entities, they are usually incomplete, i.e., some relationships between entities are missing. Accordingly, extensive research has been done on predicting those missing links through learning low-dimensional embedding representations of entities and relations~\cite{bordes2011learning,bordes2013translating,yang2014embedding,krompass2015type,nickel2015review,nguyen2016stranse,feng2016knowledge,trouillon2016complex,das2016chains,cai2017kbgan,xie2017interpretable,dettmers2018convolutional,lacroix2018canonical,ebisu2018toruse,schlichtkrull2018modeling,sun2019rotate}. Considering the large size of knowledge graphs with millions of facts, current popular link predictors tend to be fast and shallow, utilizing simple scoring functions and small embedding sizes, but at the potential expense of learning less expressive features. 


In this work, different from the majority of prior studies, the goals of which are to design new scoring functions, such as TransE~\cite{bordes2013translating}, DistMult~\cite{yang2014embedding}, ComplEx~\cite{trouillon2016complex}, RotatE~\cite{sun2019rotate}, etc, we  propose a general and effective method which could be applied to these models to boost their performance without explicitly increasing the embedding size or changing the scoring function. In more details, the original embeddings of entities and relations will be mapped to a more expressive and robust space by decompressing functions, then these link prediction models will be trained in this new space. Our method is simple and general enough to be applied to existing link prediction models and experimental results on different benchmark knowledge graphs and popular link prediction models demonstrate that our method could boost the scores with just a small amount of extra parameters.

Specifically, our contributions are as follows:
\begin{itemize}
    \item We propose \textbf{DeCom}, a simple but effective decompressing method to significantly improve the performance of many existing knowledge graph embedding models.
    \item Without the need of increasing the embedding size explicitly, DeCom is able to help link prediction models save a lot of storing space and GPU memory.
    \item By employing convolutional neural network as the decompressing network, DeCom-based models only add a few extra parameters to original models, thus being highly parameter-efficient. Even if we use the fully connected network as the decompressing network, the number of extra parameters will still be constant and not growing with knowledge graph size.
    \item Experiments on several benchmark knowledge graphs and link prediction models show the effectiveness of our model; among them, RESCAL + DeCom achieves state-of-the-art result on FB15k-237 across all evaluation metrics.
\end{itemize}
\begin{table*}[t]
    \centering
    \begin{tabular}{|c|c|c|}
        \toprule
        \textbf{Model} &  \textbf{Scoring Function}  & \textbf{DeCom Scoring Function}\\
        \midrule \midrule
        RESCAL~\cite{nickel2011three} & $\mathbf{e_{s}}^\textrm{T}\mathbf{W_{r}}\mathbf{e_{o}}$ & $f_s(\mathbf{e_{s}})^\textrm{T}f_r(\mathbf{W_{r}})f_o(\mathbf{e_{o}})$\\
         \midrule
        DistMult~\cite{yang2014embedding} & \big \langle $\mathbf{e_{s}, e_{r}, e_{o}}$\big \rangle & \big \langle $f_{s}(\mathbf{e_{s}}), f_{r}(\mathbf{e_{r}}), f_{o}(\mathbf{e_{o}})$\big \rangle\\
        \midrule 
        ComplEx~\cite{trouillon2016complex} & Re(\big \langle$\mathbf{e_{s}}$, $\mathbf{e_{r}}$, $\mathbf{e_{o}}$\big \rangle) & Re(\big \langle $f_{s}(\mathbf{e_{s}}), f_{r}(\mathbf{e_{r}}), f_{o}(\mathbf{e_{o}})$\big \rangle)\\
        \bottomrule
    \end{tabular}
    \caption{Scoring function $\mathbf{g(s, r, o)}$ of some link prediction models w/ and w/o decompressing ($\textbf{DeCom}$) layer, where $ \big \langle \cdot \big \rangle$ denotes the generalized dot product, $f_{s}, f_{r}, f_{o}$ represents decompressing operations for the subject entity, relation and object entity respectively and $\mathbf{e_{s}}$, $\mathbf{e_{s}}$ and $\mathbf{e_{s}}$ represent the embedding of the subject entity, relation and object entity respectively. 
    }
    \label{tab:score_funcs}
\end{table*}

\begin{figure}[ht]
    \centering
    \includegraphics[width=0.5\textwidth]{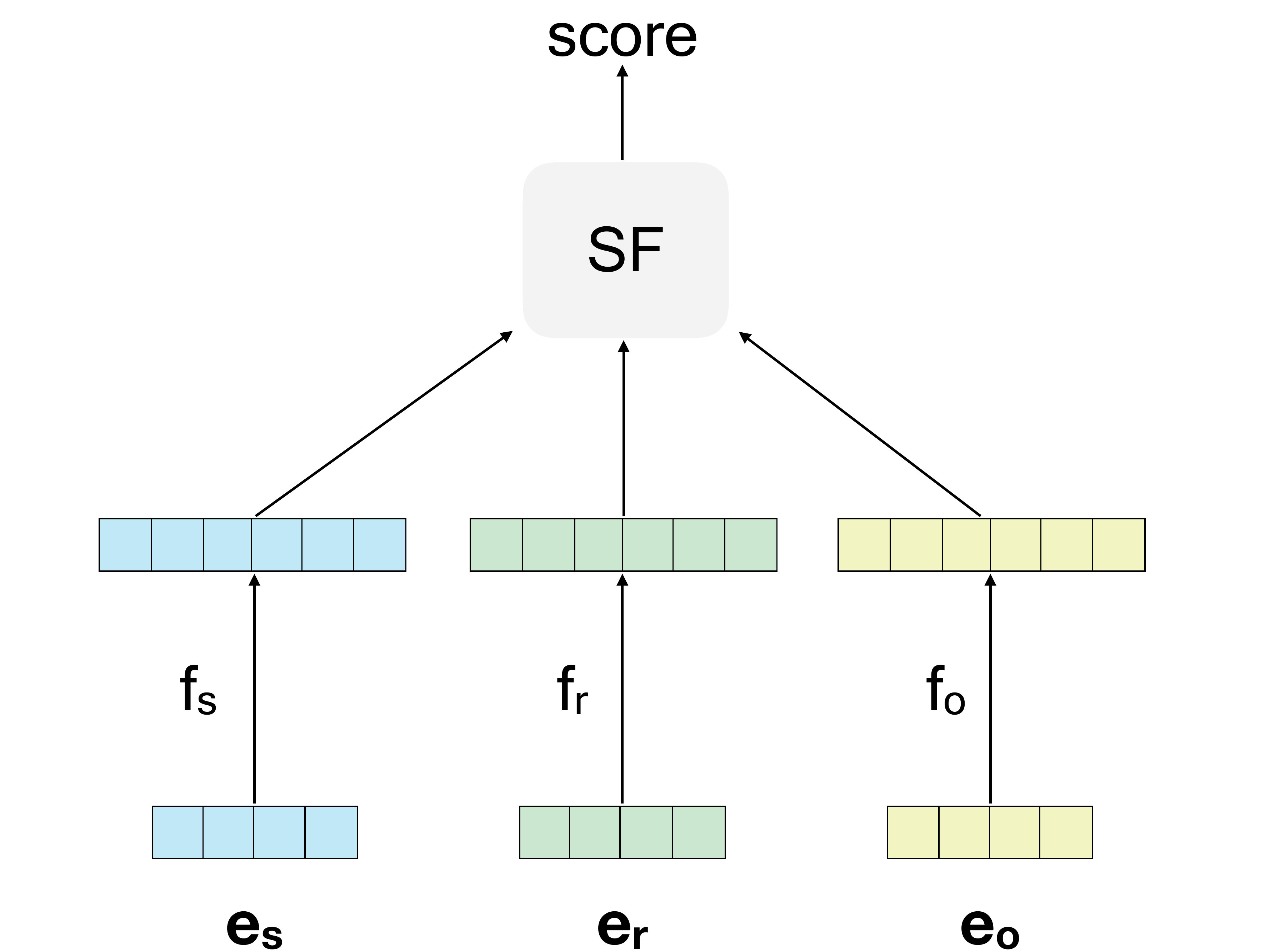}
    \caption{The architecture of DeCom-based link predicting models. Embeddings of subject, object and relations, $\mathbf{e_s}$, $\mathbf{e_o}$, $\mathbf{e_r}$ will be first fed into their corresponding DeCom layers to obtain more expressive and robust features, then the original scoring function (SF) will compute a score based on these decompressing features. The final score represents the probability of the input triple (s, r, o) being true.}
    \label{fig:Decom-graph}
\end{figure}

\section{Background}
Formally, a knowledge graph  $\mathcal{G}$ consists of entities (vertices) $\mathcal{E}$ and relations (edges) $\mathcal{R}$, and could be represented by triples (facts) \{($\mathbf{s}$, $\mathbf{r}$, $\mathbf{o}$)\} $\subseteq$ $\mathcal{E}\times\mathcal{R}\times\mathcal{E}$. Each fact (triple) represent a relationship  $\mathbf{r}\in\mathcal{R}$ between one subject entity $\mathbf{s}\in\mathcal{E}$ and  one object entity $\mathbf{o\in\mathcal{E}}$. Commonly, there are millions of entities in one knowledge graph but a lot of links (relationships) between them are missing. Therefore, completing these missing links is referred as Knowledge Base Completion (KBC), or more specifically, Link Prediction. 

Most literature approaches the link prediction task by learning low-dimensional embedding vectors of knowledge graph entities and relations, known as Knowledge Graph Embedding (KGE). They formalize the problem into finding a scoring function $g$ : $\mathcal{E}\times\mathcal{R}\times\mathcal{E}\rightarrow\mathbb{R}$, which is able to compute a score of each triple ($\mathbf{s}$, $\mathbf{r}$, $\mathbf{o}$), indicating whether this triple should be true or false. Intuitively, a promising scoring function should be able to assign higher scores to true triples than false ones. Within some of the recent models, a non-linearity such as the logistic sigmoid function is applied to the scoring function to give a corresponding probability prediction. 

Table \ref{tab:score_funcs} lists several popular scoring functions $g$ from the literature. In these models,  entities and relations are represented by low-dimensional embedding vectors, except for RESCAL where the relations are represented by full-rank matrices. 

\paragraph{RESCAL} RESCAL~\cite{nickel2011three} is a powerful link prediction model, the scoring function of which is a bilinear product between subject and object entities' embeddings and a full rank matrix for each relation. Due to its large number of parameters, RESCAL suffers from overfitting issue and explicitly increasing the relation embedding dimension will quadratically boost the number of its parameters. 

\paragraph{DistMult} 
In order to mitigate the above issue, DistMult~\cite{yang2014embedding}, a special case of RESCAL, employs a diagonal matrix to represent each relation so that the number of parameters grows linearly in terms of the embedding size. The resulting scoring function is equivalent to the inner product of three vectors. However, DistMult could not handle asymmetric relations, as ($\mathbf{s}$, $\mathbf{r}$, $\mathbf{o}$) and ($\mathbf{o}$, $\mathbf{r}$, $\mathbf{s}$) will be assigned to the same score.

\paragraph{ComplEx} To model asymmetric relations, ComplEx~\cite{trouillon2016complex} extends DistMult from the real space to the complex space. Even though each relation
matrix of ComplEx is still diagonal, the subject and object entity embeddings for the same entity are no longer equivalent, but complex conjugates, which introduces asymmetry into the tensor decomposition and thus enables ComplEx to model asymmetric relations.

\paragraph{Trade off between parameter growth and model performance}
Due to the large number of entities (vertices) in a knowledge graph, the number of parameters and computational costs are two essential aspects to evaluate a link prediction model. Specifically, since the number of entity and relation embedding parameters are $|\mathcal{E}| \times D_{e}$ and $|\mathcal{R}| \times D_{r}$, where $D_{e}$ and $D_{r}$ are the entity and relation embedding dimension respectively, a large embedding size will lead to a unmanageable number of parameters. For example, applying DistMult with an embedding size of 400 to the whole Freebase needs more than 100 GB memory to store its parameters.

As a consequence, from Table \ref{tab:score_funcs}, it is easy to find that all these popular scoring functions are simple and only contain some basic operations, such as matrix multiplications and vector products, etc. Also, they tend to set the dimensionality of entities' and relations' embedding size relatively low (around 200). As a result, these simple, small and fast models could be applicable in real-world scenarios. 
The drawback is that these relatively low-dimensional embeddings may not have the capacity to model the semantic of the knowledge graph very well (high bias), posing a negative effect on model performance.

In all, the choice of knowledge graph embedding size, though rarely discussed, is an important problem to be addressed.
In this work, we propose \textbf{DeCom}, a simple but effective method to decompress the low-dimensional embedding to a high dimensional space. Furthermore, DeCom has the following attractive aspects:
\begin{itemize}
    \item DeCom does not explicitly increase the embedding size so that the model is still able to be scaled in a manageable way.
    \item DeCom is able to not only implicitly increase the expressiveness but extract more robust features from the original embedding to achieve better performance as well, thus reduce the risk of overfitting.
    \item DeCom learns a more general representation through the decompressing network which could be easily incorporated into many existing link predictors.
\end{itemize}

\section{Motivation and Approach Overview}

Despite a large amount of literature that designs new scoring functions, there have been limited discussions about how large the embedding size should be. \citeauthor{sharma2018towards}~\shortcite{sharma2018towards} examine the geometry of knowledge graph embeddings and their experiment results suggest that for multiplicative methods (DistMult, ComplEx, etc.), increasing entity and relation embedding size leads to decreasing conicity (a high value of conicity would imply that the vectors lie in a narrow cone centered at origin) which might improve link prediction performance. 
It also has been proved that several bi-linear methods can be fully-expressive (i.e. there exists an assignment of values to the embeddings that accurately separates the correct triples from incorrect ones) given large enough embedding size~\cite{kazemi2018simple}. However, they also show that the upper bound of embedding size for full-expressiveness is $O(|\mathcal{E}||\mathcal{R}|)$, where $|\mathcal{E}|$ is the number of entities and $|\mathcal{R}|$ is the number of relations in knowledge graphs, which is not feasible even on a small-scale toy knowledge graph.

People may be encouraged to use larger embedding size by above observations, but it is not just modeling scalability that sets them back. In fact, as shown in our experiment results in Table~\ref{tab:main_wn18rr} (rows 6 vs. 13 and 14 vs. 21) , increasing embedding size does not guarantee better performance. The same phenomenon has been observed for word embeddings, and \citeauthor{yin2018dimensionality}~\shortcite{yin2018dimensionality} explains this phenomenon under the bias-variance tradeoff framework: larger embedding size leads to decreased bias (better reconstruct the factorized coorcurrence matrix), but increased variance (overfit to the noise in the matrix). The same analysis can be applied to knowledge graph embedding as well,
considering RESCAL, for example, what it essentially does is a tensor decomposition
$$ \mathcal{X} = ERE^T $$
where $\mathcal{X} \in \mathbb{R}^{|\mathcal{E}|\times|\mathcal{R}|\times|\mathcal{E}|}$ is the tensor that represents the training graph, $\mathcal{X}_{ijk} = +1$ if there is a relation $j$ from entity $i$ to entity $k$, and $\mathcal{X}_{ijk} = -1$ if there is none. $E \in \mathbb{R}^{|\mathcal{E}|\times d}$ is the entity embedding matrix where $E_{i\cdot}$ is the embedding vector of the $i$th entity and $d$ is the embedding size. $R \in \mathbb{R}^{d \times |\mathcal{R}|\times d}$ is the relation tensor where $R_{\cdot j\cdot}$ is the embedding matrix of the $j$ relation.
This decomposition can be lossless with large enough embedding size, but if we do obtain such an embedding, the performance on evaluation and test sets will be zero - as the training graph tensor $\mathcal{X}$ is corrupted from the true graph tensor $\hat{\mathcal{X}}$ by randomly flipping some of its entries. Moreover, for the link prediction task, we are exclusively evaluating those corrupted entries. It is popular for recent studies to prove that their model is fully expressive (unbiased with large enough $d$), but it may not be relevant to actual model performance, as the variance plays an important role here. Since DistMult is a special case of RESCAL and ComplEx generalizes DistMult to complex space, this analysis can be applied to them, and other multiplicative models, as well.

Motivated by the scalability issue and bias-variance tradeoff, we propose to use a shallow neural network to decompress low-dimensional embedding vectors to a higher-dimensional space before applying the scoring functions. The intuition is that the low-dimensional embeddings will store the compressed information about entities/relations, and the decompressing network will project this compressed representation into a higher-dimensional space which is easier for the simple scoring functions to handle, thus achieving the low bias of high dimensional embedding with much fewer parameters. On the other hand, the decompressing network must learn the general information about the knowledge graph, making it more robust to noise and have lower variance. Less number of total parameters also suggests the model is less prone to overfitting.

\section{Detailed Approach}

\begin{table*}[t]
    \centering
    \begin{tabular}{|c|l|cccc|cccc|}
    \toprule
        \multirow{2}{*}{\bf \#}&\multirow{2}{*}{~~~~~~~~~Model} & \multirow{2}{*}{$d_e$} &
        \multirow{2}{*}{$\hat{d}_e$} &
        \multirow{2}{*}{$d_r$} &
        \multirow{2}{*}{$\hat{d}_r$} &
        \multicolumn{4}{|c|}{\textbf{FB15k-237}}   \\
       &  & & & & &  MRR & H@1 & H@3&  H@10\\
         \midrule
         1 & \textbf{RESCAL} & 100 & - & $100^2$ & - & 0.255 & 0.185 & 0.278 & 0.397 \\
         2 & ~~~~\textbf{+ F-DeCom} & 100 & 400 & 200 & $400^2$ & 0.353 & 0.260 & 0.388 & 0.535\\
         3 &~~~~\textbf{+ F-DeCom-En} & 100 & 200 & $200^2$ & - & 0.349 & 0.260 & 0.381 & 0.526\\
         4 &~~~~\textbf{+ F-DeCom-Rel} & 200 & - & 200 & $200^2$ & 0.354 & 0.261 & 0.388 & 0.536 \\
         5 &~~~~\textbf{+ vanilla expansion} & 400 & - & $400^2$ & - & 0.317 & 0.233 & 0.344 & 0.483\\
         \midrule
         6 &\textbf{DistMult} & 100 & - & 100 & - &0.258 & 0.173 & 0.283 & 0.417\\
         7 &~~~~\textbf{+ C-DeCom} & 100 & 400 & 100 & 400 & 0.291 & 0.210 & 0.318 & 0.454\\
         8 &~~~~\textbf{+ F-DeCom} & 100 & 400 & 100 & 400 & 0.299 & 0.213 & 0.329 & 0.470\\
         9 &~~~~\textbf{+ F-DeCom-En} & 100 & 400 & 400 & - & 0.296 & 0.214 & 0.325 & 0.460\\
         10 &~~~~\textbf{+ F-DeCom-Rel} & 200 & - & 100 & 200 & 0.281 & 0.201 & 0.307 & 0.442\\
         11 &~~~~\textbf{+ C-DeCom-En} & 100 & 400 & 400 & - & 0.278 & 0.198 & 0.306 & 0.442\\
         12 &~~~~\textbf{+ C-DeCom-Rel} & 400 & - & 100 & 400 & 0.273 & 0.192 & 0.299 & 0.436\\
         13 &~~~~\textbf{+ vanilla expansion} & 400 & - & 400 & - & 0.269 & 0.186 & 0.291 & 0.428\\
         \midrule
         14 &\textbf{ComplEx} & 100 & - & 100 & - & 0.257 & 0.182 & 0.270& 0.426\\
         15 &~~~~\textbf{+ C-DeCom} & 100 & 400 & 100 & 400 & 0.284& 0.200& 0.313&  0.453\\
         16 &~~~~\textbf{+ F-DeCom} & 100 & 200 & 100 & 200 & 0.303 & 0.218 & 0.334 & 0.473\\
         17 &~~~~\textbf{+ F-DeCom-En} & 100 & 400 & 100 & 400 & 0.280 & 0.205 & 0.325 & 0.461\\
         18 &~~~~\textbf{+ F-DeCom-Rel}& 100 & 400 & 100 & 400 & 0.280 & 0.205 & 0.325 & 0.461\\
         19 &~~~~\textbf{+ C-DeCom-En} & 100 & 400 & 100 & 400 & 0.285 & 0.203 & 0.312 & 0.450\\
         20 &~~~~\textbf{+ C-DeCom-Rel} & 100 & 400 & 100 & 400 &0.283 & 0,199 & 0.323 & 0.453 \\
         21 &~~~~\textbf{+ vanilla expansion} & 400 & - & 400 & - & 0.267 & 0.188 & 0.292 & 0.426\\
        \bottomrule 
    \end{tabular}
    \caption{Performance of different models w/ and w/o decompressing on the testset of FB15k-237 dataset. C-DeCom and F-DeCom denotes the CNNs-based and FCNNs-based decompressing functions. *-En means only decompressing entity embeddings and *-Rel shows that only relation embeddings are decompressed. $d_e$ and $\hat{d}_{e}$ denote the dimension of entity features before and after decompressing layer. $d_r$ and $\hat{d}_{r}$ represent the dimension of relation features before and after decompressing layer. '-' denotes no decompressing layer in the model. `vanilla expansion` means explicitly increasing the embedding dimension (same notation are followed in other tables).}
    \label{tab:main_fb15k237}
\end{table*}

\begin{table*}[t]
    \centering
    \begin{tabular}{|c|l|cccc|cccc|}
    \toprule
        \bf\multirow{2}{*}{\#} & \multirow{2}{*}{~~~~~~~~~Model} & \multirow{2}{*}{$d_e$} &
        \multirow{2}{*}{$\hat{d}_e$} &
        \multirow{2}{*}{$d_r$} &
        \multirow{2}{*}{$\hat{d}_r$} &
        \multicolumn{4}{|c|}{\textbf{WN18RR}}   \\
        & & & & & &  MRR & H@1 & H@3&  H@10\\
         \midrule
         1 & \textbf{RESCAL} & 100 & - & $100^2$ & - & 0.441 & 0.417 & 0.452 & 0.487\\
         2 &~~~~\textbf{+ F-DeCom} & 100 & 200 & 100 & $200^2$ & 0.457 & 0.427 & 0.469 & 0.515\\
         3 &~~~~\textbf{+ F-DeCom-En} & 100 & 400 & $400^2$ & - & 0.451 & 0.424 & 0.464 & 0.500 \\
         4 & ~~~~\textbf{+ F-DeCom-Rel} & 200 & - & 100 & $200^2$ & 0.453 & 0.427 & 0.464 & 0.503 \\
         5 &~~~~\textbf{+ vanilla expansion} & 400 & - & $400^2$ & - & 0.436 & 0.415 & 0.444 & 0.475\\
         \midrule
         6 & \textbf{DistMult} & 100 & - & 100 & - & 0.427 & 0.381 & 0.436 & 0.487\\
         7 & ~~~~\textbf{+ C-DeCom}& 100 & 400 & 100 & 400 & 0.445 & 0.413 & 0.458 & 0.510\\
         8 & ~~~~\textbf{+ F-DeCom}& 100 & 200 & 100 & 200 & 0.450 & 0.418 & 0.461 & 0.515\\
         9 & ~~~~\textbf{+ F-DeCom-En} & 100 & 200 & 200 & - & 0.440 & 0.401 & 0.452 & 0.507\\
         10 & ~~~~\textbf{+ F-DeCom-Rel} & 200 & - & 100 & 200 & 0.442 & 0.396 & 0.449 & 0.508\\
         11 & ~~~~\textbf{+ C-DeCom-En} & 100 & 400 & 400 & - & 0.431 & 0.392 & 0.447 & 0.502\\
         12 & ~~~~\textbf{+ C-DeCom-Rel} & 400 & - & 100 & 400 & 0.440 & 0.411 & 0.450 & 0.505\\
         13 & ~~~~\textbf{+ vanilla expansion} & 400 & - & 400 & - & 0.422 & 0.380 & 0.437 & 0.482\\
         \midrule
         14 & \textbf{ComplEx} & 100 & - & 100 & - & 0.445 & 0.415 & 0.457 & 0.502\\
         15 &~~~~\textbf{+ C-DeCom} & 100 & 400 & 100 & 400 & 0.438 & 0.419 & 0.476 & 0.521\\
         16 &~~~~\textbf{+ F-DeCom} & 100 & 400 & 100 & 400 & 0.452 & 0.410 & 0.461 & 0.509\\
         17 &~~~~\textbf{+ F-DeCom-En} & 100 & 200 & 200 & - & 0.442 & 0.406 & 0.461 & 0.504\\
         18 &~~~~\textbf{+ F-DeCom-Rel} & 200 & - & 100 & 200 & 0.448 & 0.411 & 0.463 & 0.507\\
         19 &~~~~\textbf{+ C-DeCom-En} & 100 & 400 & 400 & - & 0.441 & 0.410 & 0.460 & 0.507\\
         20 &~~~~\textbf{+ C-DeCom-Rel} & 400 & - & 100 & 400 & 0.448 & 0.420 & 0.455 & 0.511\\
         21 &~~~~\textbf{+ vanilla expansion} & 400 & - & 400 & - & 0.440 & 0.411 & 0.452 & 0.510\\
        \bottomrule 
    \end{tabular}
    \caption{Performance of different models w/ and w/o decompressing on the testset of WN18RR dataset. 
    }
    \label{tab:main_wn18rr}
    \vspace{-3mm}
\end{table*}
\begin{table*}[t]
    \centering
    \begin{tabular}{|l|cccc|cccc|}
    \toprule
        \multirow{2}{*}{~~~~~~~~~Model} &  \multicolumn{4}{|c|}{\textbf{FB15k-237}} &  \multicolumn{4}{|c|}{\textbf{WN18RR}} \\
         &  MRR & H@1 & H@3&  H@10&  MRR & H@1 & H@3 & H@10\\
        \midrule
        \textbf{RESCAL} & 0.255 & 0.185 & 0.278 & 0.397 & 0.441 & 0.417 & 0.452 & 0.487\\
        ~~~~~~~ + \textbf{DeCom} & \textbf{0.353} & \textbf{0.260} & \textbf{0.388} & \textbf{0.535}  & 0.457 & 0.427 & 0.469 & 0.515\\
        \midrule
        \textbf{DistMult[$\spadesuit$]} & 0.241 & 0.155 & 0.263 & 0.419 & 0.430 & 0.390 & 0.440 & 0.490\\
        \textbf{DistMult (ours)} & 0.258 & 0.173 & 0.283 & 0.417 & 0.427 & 0.381 & 0.436 & 0.487\\
        ~~~~~~~ + \textbf{DeCom} & 0.299 & 0.213 & 0.329 & 0.470 & 0.450 & 0.418 & 0.461 & 0.515 \\
        \midrule
        \textbf{ComplEx[$\spadesuit$]} & 0.247 & 0.158 & 0.275 & 0.428 & 0.440 & 0.410 & 0.460 & 0.510\\
        \textbf{ComplEx (ours)} & 0.257 & 0.182 & 0.270& 0.426& 0.445 & 0.415 & 0.457 & 0.502\\
        ~~~~~~~ + \textbf{DeCom} & 0.303 & 0.218 & 0.334 & 0.473& 0.452 & 0.410 & 0.461 & 0.509\\
        \midrule
        \textbf{RotatE[$\heartsuit$]} & 0.332 & 0.235 & 0.368 & 0.524 & \textbf{0.475} & \textbf{0.433} & \textbf{0.494} & \textbf{0.556}\\
         ~~~~~ - \textbf{adv sample} & 0.297 & 0.205 & 0.328 & 0.480 &- &- &- &-\\
        \midrule
        \textbf{ConvE}[$\spadesuit$] & 0.325 & 0.237 & 0.356 & 0.501 & 0.430 & 0.400 & 0.440 & 0.520\\
        \bottomrule
    \end{tabular}
    \caption{Performance of different models w/ and w/o decompressing on the testset of FB15k-237 and WN18RR datasets. Results of [$\spadesuit$] and [$\heartsuit$] are taken from~\citeauthor{dettmers2018convolutional}~\shortcite{dettmers2018convolutional} and~\citeauthor{sun2019rotate}~\shortcite{sun2019rotate}. \textbf{-adv sample} stands for RotatE without adversarial sampling, which should be a more fair comparison. For each DeCom-enhanced result, the best result is selected from all DeCom settings from Tables~\ref{tab:main_fb15k237} and \ref{tab:main_wn18rr}.}
    \label{tab:main_res}
    \vspace{-3mm}
\end{table*}

 \subsection{DeCom}
We denote the decompressing functions as $f_{e_{s}}(\cdot)$, $f_r(\cdot)$, $f_{e_{o}}(\cdot) \in R^d \rightarrow R^D$ for the subject entity $\mathbf{{e_{s}}}$, relation \textbf{r} and objective entity $\mathbf{{e_{o}}}$ respectively, where $d$ and $D$ are the original and projected embedding sizes repsectively. For any scoring function $g(\mathbf{h}, \mathbf{r}, \mathbf{t})$, we could simply incorportate our decompressing function and change it into $g(f_h(\mathbf{h}), f_r(\mathbf{r}), f_t(\mathbf{t}))$. For example, the scoring function of DistMult is $\langle \mathbf{h}, \mathbf{r}, \mathbf{t}\rangle$, and after inserting the decompressing layer, we can change it into $\langle f_h(\mathbf{h}), f_r(\mathbf{r}), f_t(\mathbf{t}\rangle)$. Table~\ref{tab:score_funcs} shows more examples about scoring functions w/ and w/o decomressing operations. Figure \ref{fig:Decom-graph} shows the DeCom-based knowledge graph embedding model architecture. 

\subsection{Decompressing Functions}

Theoretically, DeCom could be implemented by any kinds of architectures, such as fully-connected, convolutional and recurrent neural networks. 、 In this work, we mainly explore decompressing functions via convolutional neural networks (CNNs) and fully connected neural networks (FCNNs). 
Because the embedding has no sequential nature, the recurrent neural network has not been explored in this work. 
Furthermore, we need to point out that decompressing functions,  $f_{e_{s}}(\cdot)$, $f_r(\cdot)$, $f_{e_{o}}(\cdot)$ are independent and do not need to be same. 

\paragraph{CNNs-based DeCom (C-DeCom)}
Because of the high parameter efficiency and fast computation speed, CNNs are suitable to represent decompressing functions.
 Details of the CNNs-based decompressing function are as follows: for a batch of triples, we first look up their embedding vectors from entity and relation embedding tables. Then we feed them into one layer of 1-D CNN followed by the batch normalization and dropout, and use the final output to train a knowledge embedding model. Here, Batch normalization~\cite{ioffe2015batch} and dropout~\cite{srivastava2014dropout} are employed to speed up training and prevent overfitting. Generally, this method could be easily incorporated into any non-parametric scoring functions. 
 \paragraph{FCNNs-based decompressing function (F-DeCom)}
  Similar to C-DeCom, F-DeCom employs a linear layer to decompress the input features into a higher dimensional feature space.

\section{Experiments}
We experiment with three bi-linear knowledge graph embedding models, i.e., \textbf{RESCAL}~\cite{nickel2011three}, \textbf{DistMult}~\cite{yang2014embedding} and \textbf{ComplEx}~\cite{trouillon2016complex} on two benchmark datasets, i.e., \textbf{FB15k-237}~\cite{toutanova2015observed} and \textbf{WN18RR}~\cite{dettmers2018convolutional}, to show that our proposed method could consistently boost the performance of knowledge graph embedding methods.
\subsection{Experimental Settings}
\paragraph{Benchmark Datasets.}
FB15k~\cite{bordes2013translating} and WN18~\cite{bordes2013translating} are widely used for evaluating knowledge graph embedding methods. ~\citeauthor{toutanova2015observed}~\shortcite{toutanova2015observed} shows that FB15k contains a large number of inverse relations and most test triples can be inferred from its reverse relation in the training set, so they delete the reverse relations from FB15k and propose FB15k-237. There are 14,541 entities and 237 kinds of relations in FB15k-237. Similarly, \citeauthor{dettmers2018convolutional}~\shortcite{dettmers2018convolutional} removes the reverse relations from WN18 and propose WN18RR. Therefore, in this paper, we evaluate our methods on FB15k-237 and WN18RR. There are 40,934 entities and 11 types of relations in WN18RR.
\paragraph{Evaluation Protocol.} We follow the standard evaluation protocol of this task. For each test triple $(\textbf{s}, \textbf{r}, \textbf{o})$, we corrupt subjecst or the objects in the knowledge graph into $(\textbf{s}, \textbf{r}, \textbf{o}^{'})$ or $(\textbf{s}^{'}, \textbf{r}, \textbf{o})$. Then we rank the triples and see how good the ground truth is ranked. Triples that are different from the ground truth but are also correct are filtered.
Mean Reciprocal Rank (MRR) and Hit@N (H@N) where N$\in\{1, 3, 10\}$, are standard evaluation measures
for these datasets and are reported in our experiments.
\paragraph{Different DeCom strategies}
In order to further explore DeCom, various decompressing strategies are explored and the details are the following :
\begin{itemize}
    \item Different Decompressing functions: in this work, two decompressing functions, F-DeCom and C-DeCom, are explored in our experiments.
    \item Decompressing objects: Because of the high flexibility of DeCom, decompressing functions could be applied on 1) just entities 2) just relations and 3) both entities and relations.
\end{itemize}
\paragraph{Hyperparamerter Settings.}
In order to make our results comparable, for each link predicting baseline model, we keep most of the hyperparameters and training strategies the same between the original model and DeCom-enhanced model. All models are trained for 500 epochs, embedding size is 100, and other hyperparameters are chosen based on the performance on the validation set by grid search.

For \textbf{DistMult}~\cite{yang2014embedding} and \textbf{ComplEx}~\cite{trouillon2016complex}, following \citeauthor{dettmers2018convolutional}~\shortcite{dettmers2018convolutional}, 1-1 training strategy is employed, and Adagrad~\cite{duchi2011adaptive} is used as the optimizer; besides, we regularize these two models by forcing their entity embeddings to have a L2 norm of 1 after parameter updating and the pairwise margin-based ranking loss (margin=1.0)~\cite{bordes2013translating} is employed. Furthermore, we find that regularizing entity embeddings after the decompressing layer to have a L2 norm of 1 could effectively prevent overfitting and make the training process stable. The range of the learning rate of Adagrad is \{0.08, 0.10, 0.12\}.

For \textbf{RESCAL}~\cite{nickel2011three}, we apply 1-N~\cite{dettmers2018convolutional} training strategy, employ Adam~\cite{kingma2014adam} as the optimizer and set binary cross entropy as the loss function. The range of the learning rate of Adam is \{0.01, 0.005, 0.001, 0.0005\}. Because RESCAL's relations are represented as full-rank matrices, and it's not intuitive to decompress a low-dimensional vector into a matrix by convolution, we only experiment it with fully connected networks.

For each model's corresponding DeCom-enhanced model, in order to make them comparable, the training strategies such as the optimizer, 1-1 or 1-N training, hyperparameters grid search range, etc, remain the same. Besides that, hyperparameters of the decompressing function are selected via grid search according to the performance on the validation set. The ranges of hyperparameters of the DeCom layer for the grid search are
set as follows: for C-DeCom, the number of kernel \{2, 3, 4\}, the size of kernel \{3, 4\}, for F-DeCom, the dimension of decompressed features are \{200, 400\}, for RESCAL relations only, pre-decompress dimension  \{100, 200, 400, 1000, 2000\}.

\subsection{Main Results}
Link prediction results on two datasets of three baseline models and their corresponding DeCom-based models are shown in Tables \ref{tab:main_fb15k237}, \ref{tab:main_wn18rr} and \ref{tab:main_res}. 

\textbf{DeCom vs. no DeCom:} DeCom-based knowledge graph embedding models outperform their corresponding baseline models significantly, which demonstrates the expressive power of DeCom. Also note that the DeCom models also outperform the baseline models with explicitly increased embedding size, indicating that they are more robust to overfitting.

\textbf{C-DeCom vs. F-DeCom:} The F-DeCom is able to generally obtain better scores but is more prone to be overfitting because from row 16 and rows 2, 8 in Tables~\ref{tab:main_fb15k237} and \ref{tab:main_wn18rr} respectively, the best F-DeCom feature size is 200 instead of 400. One reason that F-DeCom achieving higher scores is that it could extract features from all embedding dimensions but C-DeCom is only able to extract features in the range of kernels.

\textbf{DeCom-En vs. DeCom-Rel:} From related rows in Table~\ref{tab:main_fb15k237} and \ref{tab:main_wn18rr}, just decompressing relation features could obtain slightly better result. We attribute this to that modelling relation between entities is more complicated which needs more expressive and robust features from DeCom.

\textbf{DeCom vs. others} In Table \ref{tab:main_res} we collect the scores of best configurations from Table \ref{tab:main_fb15k237} and \ref{tab:main_wn18rr} and compare them with some other recent works. Especially, DeCom-based RESCAL link prediction models achieve state-of-the-art performance on the FB15k-237 dataset across all metrics.

We further note that DeCom could assist the original model to achieve higher improvement on the dataset with a larger number of relations. Specifically, link prediction models with DeCom achieve +16\% and +5\% averaged improvement on FB15k-237 and WN18RR. We attribute this to that WN18RR is simpler in structure and the original embedding already has the ability to extract meaningful features from the small number of relations. Explicitly increasing embedding size also makes baseline performance worse on WN18RR, which suggests that 100 dimensions may be enough. Therefore, models trained on FB15k-237 benefit more from DeCom.

\begin{table*}[t]
    \centering
    \begin{tabular}{|c|cc|cccc|}
    \toprule
      Model & Embed. Size & DeCom Size & MRR &  H@10 & Param. Size  & Speed (triples/sec)\\
        \midrule
        \textbf{RESCAL} & 100 & - & 0.258 & 0.402 & 6214300 ($\approx 6.2~\textrm{M})$ & 1378.03 \\
        \textbf{RESCAL} & 400 & - & 0.324 & 0.487 & 81977200 ($\approx 82.0~\textrm{M})$ & 1317.60\\
        \textbf{RESCAL + F-DeCom} & 100 & 400 & 0.356 & 0.537 & 33751300 ($\approx 33.8~\textrm{M})$ & 1229.68\\
        \midrule
        \textbf{DistMult} & 100 & - & 0.273 & 0.421 & 1501900 ($\approx 1.5~\textrm{M})$ & 1429.06\\
        \textbf{DistMult} & 400 & - & 0.284 & 0.435 & 6007600 ($\approx 6.0~\textrm{M})$ &1400.31\\
        \textbf{DistMult + C-DeCom} & 100 & 400 & 0.314 & 0.484  & 1501942 ($\approx 1.5~\textrm{M})$ & 1400.73\\
        \textbf{DistMult + F-DeCom} & 100 & 400 & 0.302 & 0.468 & 1583700 ($\approx 1.6~\textrm{M})$ & 1386.90\\
        \midrule
        \textbf{ComplEx} & 100 & - & 0.280& 0.432& 3003800 ($\approx 3.0~\textrm{M})$ & 1412.71\\
        \textbf{ComplEx} & 400 & - & 0.270& 0.428& 12015200 ($\approx 12.0~\textrm{M})$ &1310.36\\
        \textbf{ComplEx + C-DeCom} & 100 & 400& 0.288& 0.449& 3003882 ($\approx 3.0~\textrm{M})$ & 1275.31\\
        \textbf{ComplEx + F-DeCom} &  100 & 400 & 0.308 & 0.472 & 3949120 ($\approx 4.0~\textrm{M})$ & 1221.56\\
        \bottomrule
    \end{tabular}
    \caption{Comparison between models with different types of DeCom layers on the validation set of FB15k-237. The speed is calculated by the number of triples processed per second during predicting (validation) time. DeCom size means the size of features after decompressing layer.}
    \label{tab:param-time}
    \vspace{-3mm}
\end{table*}

\section{Analysis and Discussion}

\subsection{Parameter and Running Time Efficiency}
The decompressing layer is able to map the original embedding to a more expressive and robust feature space. One natural question is: what if we explicitly increase the embedding size? Therefore, we increase the embedding size from 100 to 400 to match the feature size after decompressing layer and compare them from different perspectives. The result is shown in Table~\ref{tab:param-time}. It is clear to find that models with decompressing layer not only achieve better performance, but are much more parameter efficient with a little sacrifice of  prediction speed. Especially, ComplEx with large embedding size instead harms the performance. We attribute this to the overfitting of two many parameters. Comparing with C-DeCom, F-DeCom obtains better scores with a little more parameters and slower decoding speed.

\begin{table}[!ht]
    \centering
    \begin{tabular}{|c|cccc|cc|}
    \toprule
        Model & $d_e$ & $\hat{d}_e$ & $d_r$ & $\hat{d}_r$ & MRR & H@10 \\
    \midrule
        \textbf{DistMult} & 100 & - & 100 & - & 0.273 & 0.421\\
        ~~~~~~+\textbf{C-DeCom} & 100 & 100 & 100 & 100 & 0.291 & 0.452\\
        ~~~~~~+\textbf{F-DeCom} & 100 & 100 & 100 & 100 & 0.287 & 0.448\\
    \midrule
        \textbf{ComplEx} & 100 & - & 100 & - & 0.280 & 0.432\\
        ~~~~~~+\textbf{C-DeCom} & 100 & 100 & 100 & 100& 0.283 & 0.441\\
        ~~~~~~+\textbf{F-DeCom} & 100 & 100 & 100 & 100& 0.292 & 0.458\\
    \bottomrule
    \end{tabular}
    \caption{The robustness comparison between DeCom and original models on FB15k-237 validation set. }
    \label{tab:decom_expressive}
\end{table}
\subsection{Why is DeCom effective?}
We think that there are two main reasons to explain the effectiveness of DeCom:

\noindent1) Implicitly increasing the feature dimension to improve model's expressiveness by decompressing functions.  

\noindent2) Learning more robust features. To further understand this fact, decompressing functions are designed to keep the size of input features (original embedding) and output ones the same. Specifically, we set the output feature size of F-DeCom and C-DeCom the same as the input embedding dimension, i.e., 100. The result is shown in Table~\ref{tab:decom_expressive}. 
Despite there is no increase in embedding size, the DeCom models still achieve the performance improvement, suggesting that they could learn more robust embeddings.

\section{Related Work}
In order to predict the missing links in knowledge graphs,  
knowledge graph embedding (KGE) methods have been extensively studied in recent years. 
For example, RESCAL~\cite{nickel2011three}  employs a bilinear product between vector embeddings for each subject and object entity and a full rank matrix for each relation.
TransE~\cite{bordes2013translating} implicitly models relations through representing each relation as a bijection between source and target entities. DistMult~\cite{yang2014embedding}, as a special case of RESCAL, uses a diagonal matrix for each representation so that the amount of parameters grows linearly.  
ComplEx~\cite{trouillon2016complex} extends DistMult through modeling asymmetric relations by introducing complex embeddings. RotatE~\cite{sun2019rotate} models the relation as a rotation operation from the subject entity to the object entity in the complex vector space.
Most prior methods are based on simple operations and shallow neural network, which make them fast, scalable and memory-efficient, however, these properties also restrict the expressiveness of learned features. Concurrently, in order to mitigate this problem, \citeauthor{dettmers2018convolutional}~\shortcite{dettmers2018convolutional} (ConvE) employs 2-D convolution operations on the subject entity and relation embedding vectors, after they are reshaped to matrices and concatenated. However, the reshaping and concatenation operations and applying 2-D convolution on word embeddings are not intuitive.

\section{Conclusion}
 In this work, in order to increase expressiveness and robustness of shallow link predictors, we propose, \textbf{DeCom}, a flexible decompressing mechanism which is able to map low-dimensional embeddings to a more expressive and robust space by adding just a few extra parameters.  DeCom could be easily incorporated into many existing knowledge graph embedding models and experimental results show that it could boost the performance of many popular link predictors on several knowledge graphs and obtain state-of-the-art results on FB15k-237 across all evaluation metrics.
\newpage
\bibliography{aaai20}
\bibliographystyle{aaai}
\end{document}